# Simple but Effective Unsupervised Classification for Specified Domain Images: A Case Study on Fungi Images

Zhaocong Liu[1,2], Fa Zhang[3], Lin Cheng[1,2], Huanxi Deng[1], Xiaoyan Yang[3], Zhenyu Zhang[1,2,*], and Chichun Zhou[1,2,*]

[1]School of Engineering, Dali University, Dali, Yunnan, 671003, China.

[2]Air-Space-Ground Integrated Intelligence and Big Data Application Engineering Research Center of Yunnan Provincial Department of Education, Yunnan, 671003, China.

[3]Institute of Eastern-Himalaya Biodiversity Research, Dali University, Dali 671003, China.

[*] Corresponding author: Chichun Zhou (zhouchichun@dali.edu.cn),

Zhenyu Zhang (zhangzhenyu@dali.edu.cn)

Emails of other authors: Zhaocong Liu: liuzhaocong@stu.dali.edu.cn

Fa Zhang: zhangfa@eastern-himalaya.cn

Lin Cheng: chenglin@stu.dali.edu.cn

Huanxi Deng: denghuanxi@stu.dali.edu.cn

Xiaoyan Yang: yangxiaoyan@eastern-himalaya.cn

## Keywords

# Abstract

High-quality labeled datasets are essential for deep learning. Traditional manual annotation methods are not only costly and inefficient but also pose challenges in specialized domains where expert knowledge is needed. Self-supervised methods, despite leveraging unlabeled data for feature extraction, still require hundreds or thousands of labeled instances to guide the model for effective specialized image classification. Current unsupervised learning methods offer automatic classification without prior annotation but often compromise on accuracy. As a result, efficiently procuring high-quality labeled datasets remains a pressing challenge for specialized domain images devoid of annotated data. Addressing this, an unsupervised classification method with three key ideas is introduced: 1) dual-step feature dimensionality reduction using a pre-trained model and manifold learning, 2) a voting mechanism from multiple clustering algorithms, and 3) post-hoc instead of prior manual annotation. This approach outperforms supervised methods in classification accuracy, as demonstrated with fungal image data, achieving 94.1% and 96.7% on public and private datasets respectively. The proposed unsupervised classification method reduces dependency on pre-annotated datasets, enabling a closed-loop for data classification. The simplicity and ease of use of this method will also bring convenience to researchers in various fields in building datasets, promoting AI applications for images in specialized domains.

# 1. Introduction

High-quality annotated datasets are essential for fully unlocking the capabilities of deep learning. Supervised deep learning models, powered by vast datasets, such as ImageNet (Deng et al., 2009), COCO (Lin et al., 2014), and Open Images Dataset V4 (Kuznetsova et al., 2020), have achieved human-parity accuracy in generic image classification (Simonyan & Zisserman, 2014; Tan, & Le, 2019). Furthermore, they've realized substantial progress in image segmentation (Long, Shelhamer, & Darrell, 2015; Carion et al., 2020) and generation (Goodfellow et al., 2014; Karras et al., 2020).

While generic images capture scenes frequently encountered in daily life, specialized domain images delve into rarer subjects, encompassing astronomical images (York et al., 2000; Zhou et al., 2022), fungal depictions (Luo et al., 2019), and images of aquatic plants (Madsen & Wersal 2017). Moreover, specialized images often involve captures from professional tools, including medical X-ray machines (Huda & Abrahams, 2015), ultrasound devices (Liu et al, 2019), and hyperspectral imaging equipment (Zhang & Du, 2012), as shown in Figure 1. Spurred by successes in the generic domain, numerous specialized fields are now fervently gathering and annotating data (Litjens et al., 2017; Zhou et al., 2022; Fang et al., 2023; Dai et al., 2023), setting a foundational base for AI implementation across varied vertical domains.

Traditional manual data annotation is both costly and time-consuming. Moreover, annotating domain-specific data often demands domain expertise, making efficient and accurate annotation even more challenging. For instance, the generic image dataset, ImageNet, required a large number of annotators to invest years to achieve precise annotations (Deng et al., 2009). Similarly, in the astronomy field, the GalaxyZoo database depended on hundreds of thousands of volunteers, taking months to finalize the annotations (Lintott et al., 2008). Our previous work (Zhou et al., 2022; Fang et al., 2023) showed that the GalaxyZoo dataset exhibited subjective biases due to the annotators' lack of domain expertise. While manual annotation has been pivotal for image recognition, its high costs and limited scalability render it impractical for other specialized domain image classification. Consequently, specialized domain images starved of labeled data are in dire need of innovative automatic annotation methodologies.

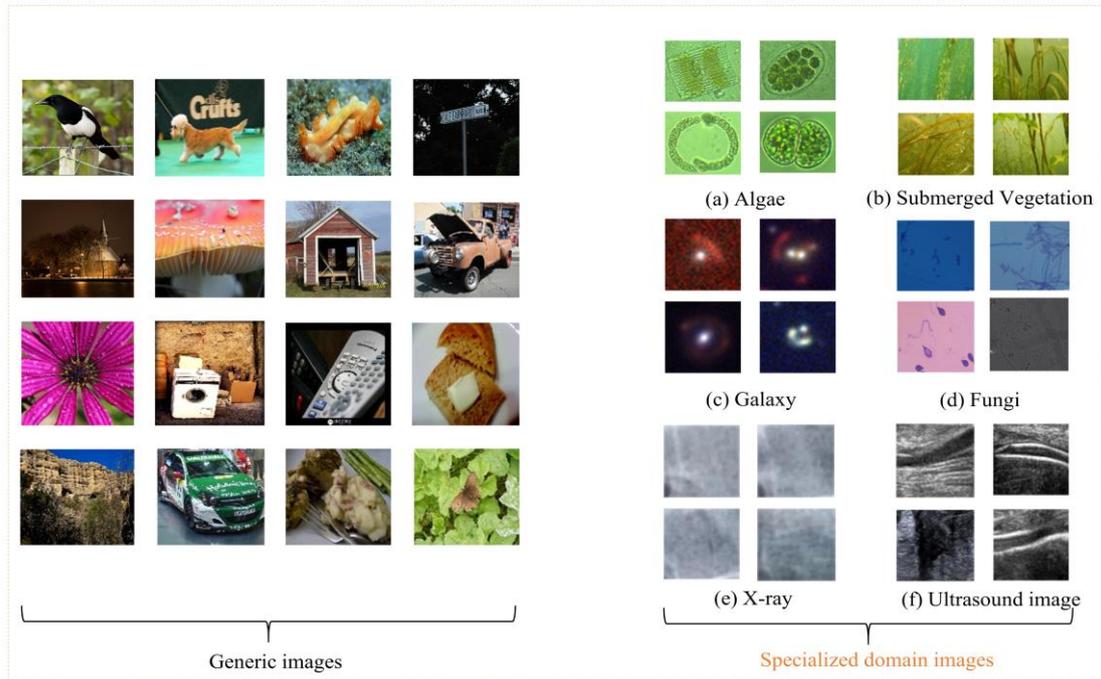

Figure 1．Comparison between generic domain images and specialized domain images. Algae images are sourced from EMDS-7 (Yang et al., 2023), galaxy images from DESI (Dey et al., 2019) fungal and aquatic plants images from a private dataset, and X-ray and ultrasound images from the internet.

In traditional supervised learning, both the feature extraction module, such as convolutional neural networks (CNN) (Conneau et al., 2016), and the classifier, such as multi-layer perception (MLP) (Zhou et al., 2020; Kruse et al., 2022), are often trained concurrently. However, this joint training demands a vast and highly accurate annotated dataset to guide the model in discerning effective features from a myriad of possibilities and to understand the correlation between these features and labels. For instance, the previous work (Fang et al., 2023) highlighted that to identify galaxy rotation features, a training set comprising images across a broad range of angles is crucial. In specialized domains where annotated data is scarce, neither the feature extraction nor the classification modules receive optimal training. This insufficiency hampers the performance of supervised models. In response to this shortfall of annotated data in specialized domains, researchers are exploring alternative training strategies: namely, self-supervised learning-based (Jing & Tian, 2020; Ohri & Kumar, 2021; Kumar, Rawat, & Chauhan, 2022; Ericsson et al., 2022; Ozbulak et al., 2023) classification and unsupervised learning-based (Hou et al., 2021; Schmarje et al., 2021; Zhou et al., 2022; Zhu et al., 2022; Ozbulak et al., 2023) classification.

**Self-supervised learning-based classification methods.** The fundamental premise of this approach is the segregation of feature extraction from classifier training. Research indicates that feature extraction doesn't necessarily require manual labeling for guidance (Jing & Tian, 2020; Ohri & Kumar, 2021; Kumar, Rawat, & Chauhan, 2022; Ericsson et al., 2022; Zhou et al., 2022; Zhou et al., 2023). Pretrained modules on extensive image datasets, such as vision transformers (VIT) (Paul & Chen, 2022) and ConvNeXt (Liu et al., 2022), can adeptly extract features from analogous domains. Furthermore, frameworks like deep clustering (Ajay et al., 2022), such as SCAN (Van Gansbeke et al., 2020), SWAV (Caron et al., 2020) and contrastive learning techniques (Le-Khac, Healy, & Smeaton, 2020; Albelwi, 2022), such as SimCLR (Chen et al., 2020), enhance feature extraction by emphasizing differences between samples, eliminating the need for data labels. Without labeled data, self-supervised learning models derive data features from intrinsic sample signals. For downstream tasks, as the feature extraction burden is lessened, the model simply focuses on mapping features to true labels, diminishing the reliance on manual annotations. In highly similar domains, they can even venture into few-shot or zero-shot learning (Rahman, Khan, & Porikli, 2018; Kadam & Vaidya, 2020; Chen et al., 2021)

**Unsupervised learning-based classification methods.** The unsupervised methods adopt a similar feature extraction philosophy as self-supervised techniques. For instance, autoencoders (Hou et al., 2021), transfer learning (Zhou et al., 2023), and contrastive learning strategies (Chen et al., 2020; Gao et al., 2023) can be employed to extract features from data. Their distinct trait lies in utilizing clustering methods for automatic data classification (Hou et al., 2021; Zhu et al., 2022; Zhou et al., 2022). Hence, many discourses don't sharply differentiate between self-supervised and unsupervised learning methods (Ozbulak et al., 2023).

While existing self-supervised techniques have significantly alleviated the reliance on pre-labeled datasets, they're not without limitations. Domain biases between specialized and generic domains, owing to differences in data collection environments (Zhang et al., 2022) and task types (Niu et al., 2020; Liu et al., 2021a), pose challenges. Contrary to the expectation of minimal manual intervention, self-supervised learning often demands hundreds or even thousands of manually labeled samples to aptly guide the model in learning the feature-label correlation. This is evident as increasing the number of training sets markedly boosts accuracy in few-shot learning based on self-supervision (Chen et al., 2021). When trained with minimal labeled samples, self-supervised learning's accuracy falters, trailing behind traditional supervised methods. Though unsupervised techniques can entirely sidestep the dependency on pre-labeled datasets, they usually marred by their subpar accuracy, especially when juxtaposed against supervised methods. For instance, the accuracy of individual clustering methods leaves much to be desired (Hou et al., 2021; Zhou et al., 2022). Moreover, aligning clustered categories with actual categories can be labor-intensive if clustering accuracy is off the mark (Zhou et al., 2022).

In essence, achieving precise and cost-effective labeling for specialized domain images remains a challenge.

This paper presents an innovative unsupervised classification approach, distinct in three fundamental ways from existing methods:

1) Dual-step dimensionality reduction: The method employs a two-pronged strategy for reducing feature dimensionality. Initially, it harnesses a pretrained large model to extract pertinent features within single image pixels. Owing to their training on vast datasets and intricate hierarchies, such pretrained models (Paul & Chen, 2022; Liu et al., 2022) offer multi-layered feature representations, enabling a more generalized image expression. Subsequently, manifold learning techniques (Izenman, 2012) pare down the dimensionality between samples. These nonlinear dimensionality reduction tools discern and capture intricate nonlinear data relationships, eradicating superfluous features and accentuating sample differences, as shown in Figure 2-a.

2) Bagging based clustering voting mechanism (see previous works Hou et al.,

(2021), Zhou et al., (2022), Zhu et al., (2022), Fang et al., (2023), and Dai et al., (2023)): To counteract the accuracy limitations of singular clustering approaches, our method amalgamates results from diverse clustering methodologies. This ensemble perspective enhances the robustness of the analysis. Even though this mechanism occasionally dismisses certain samples, it effectively rectifies the accuracy shortcomings inherent to individual clustering models.

3) Post-hoc label alignment: In a departure from traditional practices, our method opts for manual post-training label alignment, sidestepping the need for pre-training manual labels. Specifically, we generate clustering outcomes with an augmented category count, subsequently subjecting them to visual categorization. For instance, in a 4-class classification, the approach may yield up to twenty categories. While post-clustering manual label alignment is requisite, the enhanced accuracy means annotating twenty clusters is far more efficient than pre-labelling hundreds or thousands of individual samples, as shown in Figure 2-b. Collectively, with these advancements, our approach promises high-precision classification for specialized domain data while significantly minimizing the need for manual annotation.

Drawing from the aforementioned principles, simple and out-of-the-box techniques are integrated to craft our method.

1) Pretrained large model, ConvNeXt (Liu et al., 2022) is used to act as the feature extractor for single image sample.

2) The uniform manifold approximation and projection (UMAP) method (McInnes, Healy, & Melville, 2018) is utilized to achieve nonlinear reduction of dimensions across all samples.

3) Three distinct clustering algorithms: K-Means (Hartigan & Wong, 1979), Balanced Iterative Reducing and Clustering using Hierarchies (Birch) (Zhang, Ramakrishnan, & Livny, 1996), and the Agglomerative Clustering algorithm (Agg) (Ackermann et al., 2014) are incorporated through a voting mechanism (Hou et al., 2021; Zhou et al., 2022; Zhu et al., 2022; Fang et al., 2023; Dai et al., 2023) to give the final classifications.

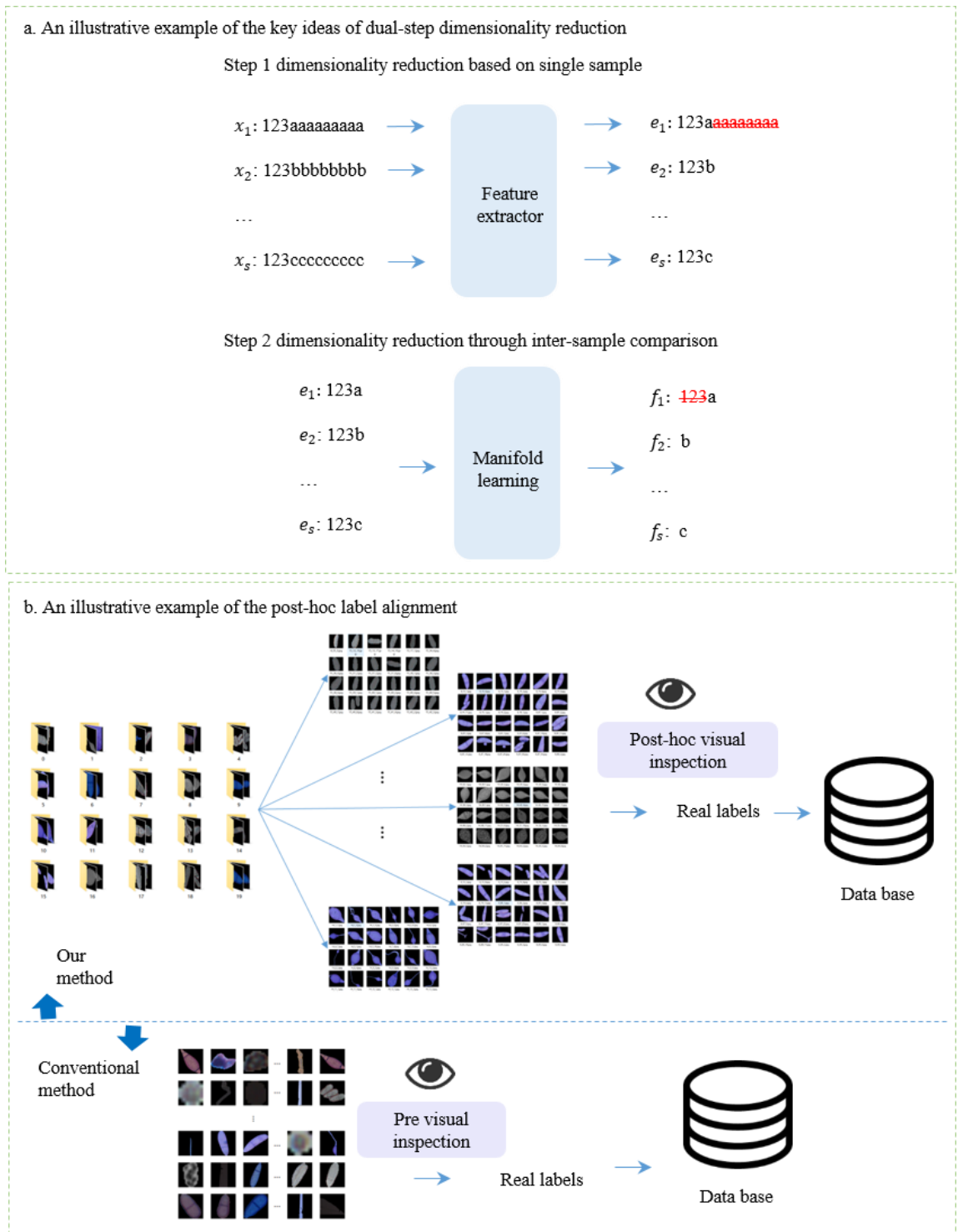

Figure 2. Illustration of the main idea of dual-step dimension reduction (panel-a) and post-hoc visual inspection for label alignment (panel-b). In the dual-step dimension reduction, the first step removes redundant information inherent in the data, retaining only the effective features. The second step, by contrasting differences between samples, further eliminates redundant features and accentuates the distinctive ones.

Fungal image data serves as an exemplary specialized domain dataset, which we've chosen to evaluate the efficacy of our proposed method. Our results are promising: on public fungal image datasets, our approach achieves a classification accuracy of 94.1%, surpassing current supervised methods with accuracy 86.0%. The method's prowess extends to private fungal image datasets as well, where it registered a 96.7% classification accuracy on over 1,000 images in a mere 10 minutes, encompassing both automatic clustering and manual post-labeling.

This underscores our unsupervised method's capacity for swift and accurate classification within specialized domains. Another significant insight is the potential reuse of datasets labeled through our unsupervised method to train supervised models. Given the absence of bias between training and testing datasets and the label accuracy rivaling expert annotation, this paves the way for creating efficient online supervised classification models. To conclude, the main contribution of the present paper can be summarized as follows:

1) The proposed unsupervised classification method reduces dependency on pre-annotated datasets, enabling a closed-loop for data classification.

2) The simplicity and ease of use of this method will also bring convenience to researchers in various fields in building datasets, promoting AI applications for images in specialized domains.

## 2. Methods, materials, and preprocess

### 2.1 Methods

The method presented in this article integrates three core techniques derived from existing out-of-box methods, further complemented by a post-hoc label alignment approach. While each technique has its roots in established technologies, their combined application is designed to comprehensively extract features from image data. This synergy strips away superfluous information, maintaining only distinct data, thus enabling high-precision classification.

1) ConvNeXt (Liu et al., 2022) is an enhancement over the ResNet (He et al., 2016) convolutional network, incorporating training techniques similar to the Swin Transformer (Liu et al., 2021b). It replaces the main architecture of ResNet with a patchify layer and broadens the model's network width through convolutional grouping. This design amalgamates the strengths of convolutional networks with Transformers, inheriting the powerful feature extraction capabilities of CNNs while introducing self-attention mechanisms to model long-term dependencies. ConvNeXt has demonstrated superior performance in image classification tasks (Liu et al., 2022). Compared to pure Transformer based models, such as VIT (Paul & Chen, 2022), ConvNeXt maintains excellent performance while significantly reducing parameter and computational overheads. As illustrated in Figure 3-a, after feature extraction by the pretrained ConvNeXt without any fine-tune, each image $x$ with resolution *256\*256* is encoded to a vector $e$ with 2048 dimensions, effectively eliminating redundant information inherent in the single image.

2) After the large model extracts features, most redundant information in the image data is removed, leaving rich, essential feature representations. However, whether a sample's feature is redundant can be determined by comparing it with other samples, see step 2 of Figure 1-a. Features present in all samples are redundant and should be discarded. While PCA (Maćkiewicz & Ratajczak, 1993) is the simplest method to find sample differences, in this study, we adopted the nonlinear dimensionality reduction method UMAP (McInnes, Healy, & Melville, 2018), which will be later prove to be more effective than PCA. UMAP, grounded in manifold learning and topological concepts, aims to preserve both local and global data structures for better data representation in lower dimensions. UMAP first calculates local similarities between each data point and its neighboring data points in high-dimensional space. By connecting points with high similarity, a minimal spanning tree is constructed in high-dimensional space. In low-dimensional space, adjacency graphs and fuzzy topological structures are similarly constructed. Optimization algorithms like stochastic gradient descent adjust the low-dimensional representation to minimize the difference between high and low-dimensional topological structures. UMAP retains both local and global data structures during dimensionality reduction, making visualization and clustering more effective. As shown in Figure 3-b, after UMAP dimensionality reduction, each image $x$ is compressed from the 2048-dimensional $e$ to the 200-dimensional $f$, further eliminating dataset-defined redundant information and retaining inter-sample differential information.

3) The bagging based multi-clustering strategy (Hou et al., 2021; Zhou et al., 2022) is an ensemble clustering method that combines the results of multiple different clustering algorithms. By examining clustering results from different perspectives, it enhances clustering accuracy and stability. Kmeans (Hartigan & Wong, 1979) divides dataset samples into k clusters using various distance formulas. Initial cluster centers are initialized by mean vectors and are updated by minimizing the distance between samples and cluster centers using a greedy strategy. Agg (Ackermann et al., 2014) and Birch (Zhang, Ramakrishnan, & Livny, 1996) are hierarchical clustering algorithms. The Agg algorithm adopts a bottom-up aggregation strategy, initially viewing each sample as an initial cluster and merging the two closest clusters at each step. The Birch algorithm performs hierarchical clustering using a Clustering Feature (CF) tree. It first constructs a CF tree based on input data, then applies clustering algorithms and outlier handling on leaf nodes. As shown in Figure 3-c, in the voting, multiple different clustering algorithms are first used to cluster the data. A voting mechanism then determines which cluster each data point ultimately belongs to. For example, using three different clustering algorithms to cluster data, pseudo-labels based on one clustering method's results are given, and each sample's category allocation in different methods is counted. For samples allocated to the same category in all three clustering methods, they are retained as the final clustering result. For some data, if the clustering category results of the three clustering algorithms are inconsistent, that data point will be discarded, termed as "rejected data". This approach enhances clustering robustness and accuracy, especially when facing noise in the data or the limitations of different algorithms.

4) Post-labeling alignment involves manually categorizing the clustering results generated by the model through visual inspection, rather than directly using manually pre-labeled samples to guide classification during training, see Figure 3-d. Specifically, the data is clustered into a number of categories several times more than needed. For example, in the case of 2-class samples, clustering results with around twenty classes are generated. Then manually categorize them based on visual inspection. Although our method still requires manual alignment of labels for clustered samples afterward, given the high clustering accuracy, post-labeling twenty high-accuracy folders is more efficient than pre-labeling hundreds or thousands of samples.

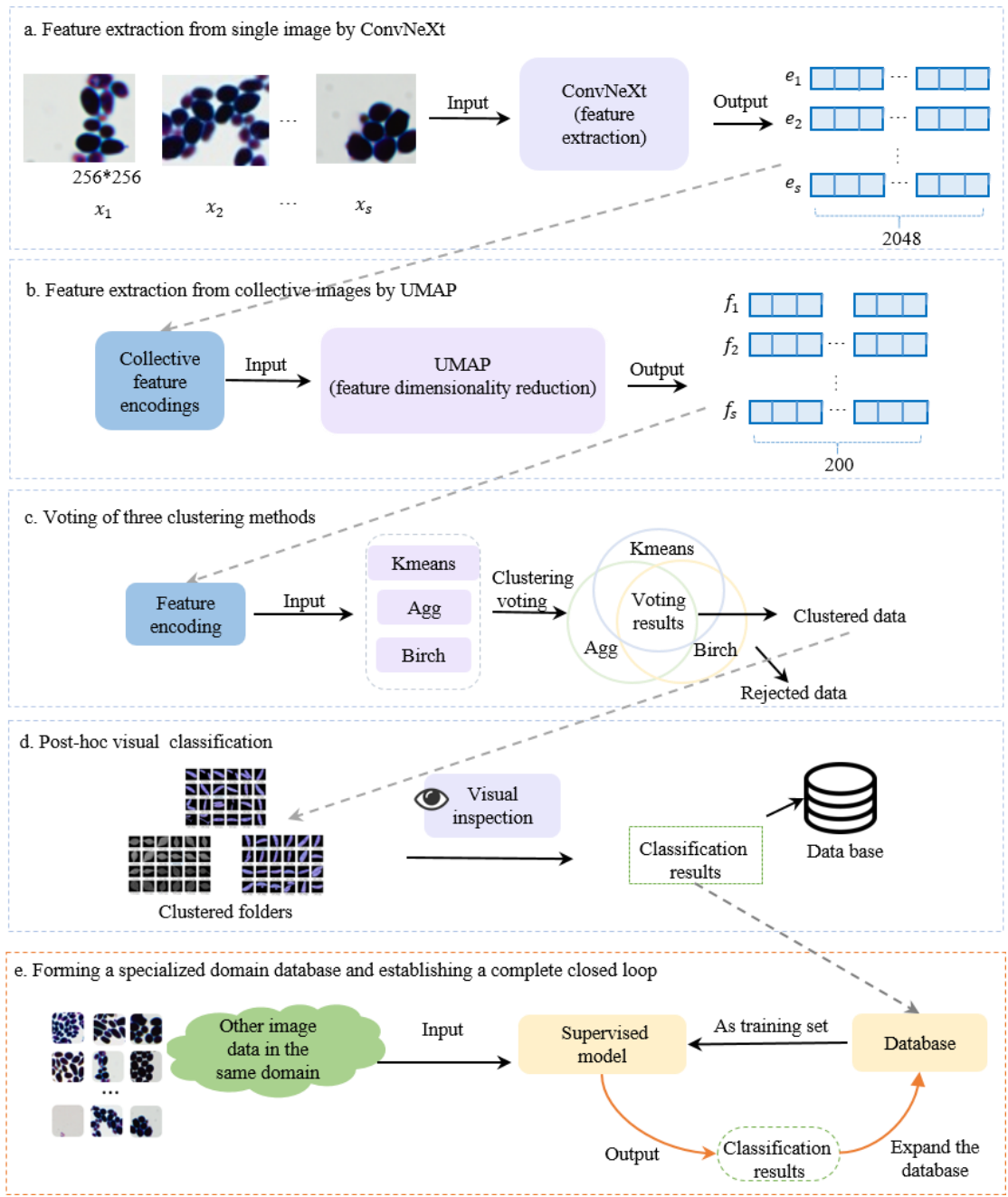

Figure 3. An overview of the main method.

In addition, to facilitate future supplements to the database, the unbiased labeled image data obtained through this classification method can be used as a training set to train an online supervised classification model, as illustrated in Figure 3-e. This process forms a closed loop for the establishment of a database, providing an effective approach for building a collaboratively evolving specialized domain database.

## 2.2 Materials

This study aims to validate the effectiveness of our method through fungal images, a representative dataset in the specialized domain. Both private and public fungal datasets are involved.

The private dataset, depicted in Figure 4-a, contains images of four species of nematode-trapping fungi from the Arthrobotrys genus: Arthrobotrys musiformis, Arthrobotrys sphaeroides, Arthrobotrys superba, and Arthrobotrys xiangyunensis. These fungi, belonging to the Ascomycota, Orbiliomycetes, Orbiliaceae families, have the unique ability to produce trapping structures to capture nematodes. Beyond their specific ecological functions and significant potential in bio-controlling harmful nematodes, these fungi have emerged as a model group for studying fungal evolution, classification, ecology, and physiology. A notable feature is their high morphological diversity, particularly in spore morphology, which aids this study (Zhang & Mo, 2006; Yang & Zhang, 2014; Zhang et al., 2023). The morphological images of the nematode-trapping fungi used in this study were captured using an Olympus BX53 microscope (Olympus Corporation, Japan). These images, taken under bright field or differential interference mode with a high-power lens, offer an in-depth view of the intricate morphological characteristics of the spores in these fungi. The creation of a dataset centered on nematode-trapping fungi holds significant value.

The DIFaS (Public Digital Image of Fungus Species) dataset (Zieliński et al., 2020) comprises images of nine different fungal species. Figure 4-b provides two random image examples from each category. The DIFaS dataset is used to assess the performance of our proposed method by comparing it with existing methods. A breakdown of the categories and quantities for the two datasets can be found in Table 1-a.

Table 1-a. The categories and quantities of the raw dataset.

| Our Dataset (Raw) | Category | Arthrobotrys musiformis (Mus) | | Arthrobotrys sphaeroides (Sph) | | Arthrobotrys superba (Sup) | | Arthrobotrys xiangyunensis (Xye) | | Total |
|---|---|---|---|---|---|---|---|---|---|---|
| | Counts | 118 | | 97 | | 68 | | 99 | | 382 |
| DIFaS (Raw) | Category | CA | CG | CL | CN | CP | CT | MF | SB | SC | Total |
| | Counts | 20 | 20 | 20 | 15 | 20 | 20 | 21 | 20 | 20 | 176 |

Table 1-b. The categories and quantities of the processed dataset.

| Our Dataset | Category | Mus | | Sph | | Sup | | Xye | | Total |
|---|---|---|---|---|---|---|---|---|---|---|
| | Counts | 324 | | 125 | | 423 | | 129 | | 1001 |
| DIFaS | Category | CA | CG | CL | CN | CP | CT | MF | SB | SC | Total |
| | Counts | 200 | 200 | 200 | 113 | 200 | 200 | 175 | 189 | 225 | 1902 |

## 2.3 Preprocess

The preprocessing helps improve image classification performance, especially when dealing with large-sized images with multi-location distributions. In the private dataset, the morphological characteristics of conidiospores can provide information about the external structure and characteristics of fungi, thus helping identify fungal types. Given the large image sizes in the dataset and the irregular spatial distributions of conidiospores, as well as susceptibility to interference factors like hyphae and bubbles during the classification process, an image segmentation model (Kirillov et al., 2023) is a solution for more effective classification. In this regard, the segment anything model (SAM) (Kirillov et al., 2023) is used to automatically segment out effective fungal spore images as well as other clutter interference images. As shown in Figure 5-a, the large fungal images are input into SAM for segmentation into several subimages containing useful conidiospore images as well as useless hyphae, bubble backgrounds and other interference images.

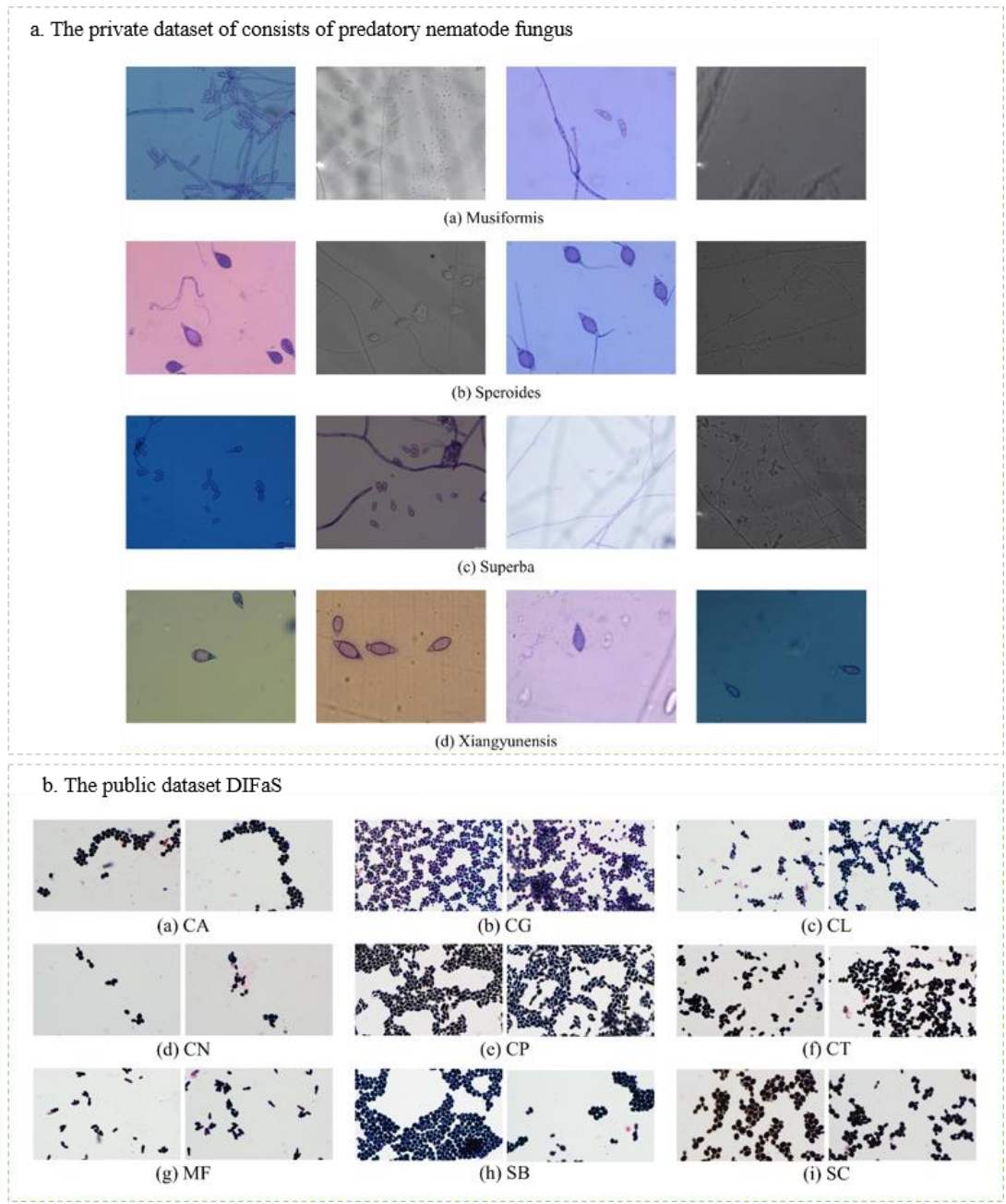

Figure 4. Examples of the fungal images. Panel-a. The public dataset DIFaS. Panel-b. The private dataset consists of predatory nematode fungus.

When processing the DIFaS dataset, we followed the standard procedures used in existing methods to ensure a fair comparison with others. Only a simple region segmentation is employed to simulate fungal images as observed by the human eye under a microscope. In the images segmented from the DIFaS dataset, backgrounds (cell-free areas) are also included, as illustrated in Figure 5-b. A breakdown of the categories and quantities for the two datasets after preprocessing can be found in Table

1-a.

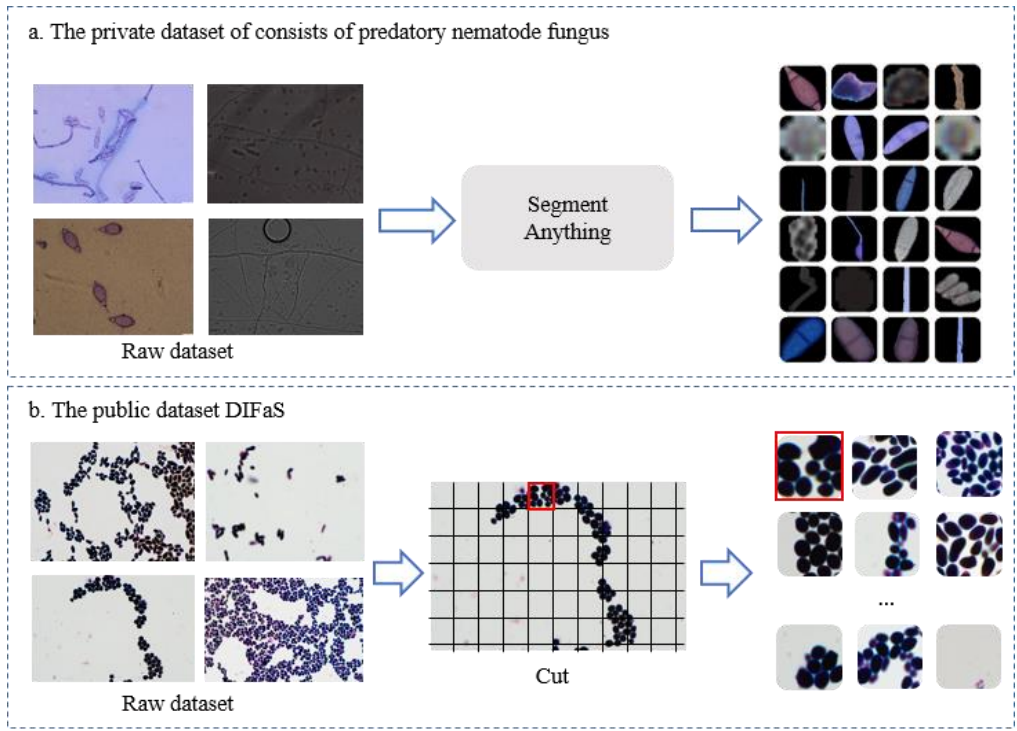

Figure 5. Illustration of the dataset preprocessing. Panel-a. The preprocess of the private dataset by using the SAM. Panel-b. The preprocess of the public dataset by cutting.

## 3. Result and analysis

### 3.1 The main result of unsupervised classification

After five repeated experiments, the approach consistently achieved an average classification accuracy of 94.1% for the DIFaS dataset and 96.7% for the private dataset, see Table 2. The confusion matrices showing the details of classifications are given in Figure 6-a. The sample discard rates generated during the voting process are 28.5% and 17.2%, respectively.

The t-SNE (Van der Maaten & Hinton, 2008) visualization is employed to analyze the outcomes at each processing stages, as depicted in Figure 6-b. Initially, the feature space learned by the ConvNeXt extractor demonstrates a clustering structure, even when compared to the original images. Subsequently, after applying nonlinear dimensionality reduction using UMAP, samples from different categories become more distinguishable in the t-SNE mapping. This suggests that UMAP amplifies the distinctiveness of sample features. Lastly, the voting process filters out numerous outliers, which aids in enhancing clustering accuracy. These visual insights underscore the combined and effective contributions of the three modules: pretrained model feature extraction, nonlinear dimensionality reduction, and ensemble classification, all of which bolster the performance of unsupervised clustering. In addition, the obtained clustering results are also displayed, as shown in Figure 7. Due to the high purity of the clustering, the pressure for later manual alignment of labels is significantly reduced.

For the DIFaS fungal image dataset, we benchmarked our unsupervised classification approach against other supervised methods, as detailed in Table 3. In the realm of classification accuracy, our unsupervised technique registers an impressive 94.1%, surpassing the existing supervised methods which have accuracies of 82.4% and 86.0%. This heightened accuracy does come with a trade-off: we had to discard about 28.5% of ambiguous samples. Notably, in the preprocessing phase, our approach relies solely on basic region cropping, eschewing intricate image enhancements. This leads to a marginally reduced sample size, with roughly 1359 samples effectively classified out of a total of 1902, in comparison to the test set used by the supervised methods. For those samples that are set aside, a subsequent supervised model can be trained for secondary predictions. Such a model can also be harnessed for real-time classification, aiding in the expansion of future databases.

Table 2. The average precision of each categories obtained by our method after five trials.

| | CA | CG | CL | CN | CP | CT | MF | SB | SC | BG | Overall (%) | Reject (%) |
|---|---|---|---|---|---|---|---|---|---|---|---|---|
| DIFaS | 96.7 ±0.4 | 100.0 ±0.0 | 93.9 ±0.4 | 73.4 ±6.1 | 99.5 ±0.3 | 89.4 ±3.9 | 95.8 ±0.2 | 98.1 ±1.8 | 80.5 ±3.2 | 99.4 ±0.1 | 94.1 ±0.6 | 28.5 ±1.8 |
| | Mus | | Sph | | Sup | | Xye | | | | Overall(%) | Reject (%) |
| Our Dataset | 95.4±0.5 | | 100.0±0.0 | | 97.4±0.2 | | 94.1±0.4 | | | | 96.7±0.2 | 17.2±2.5 |

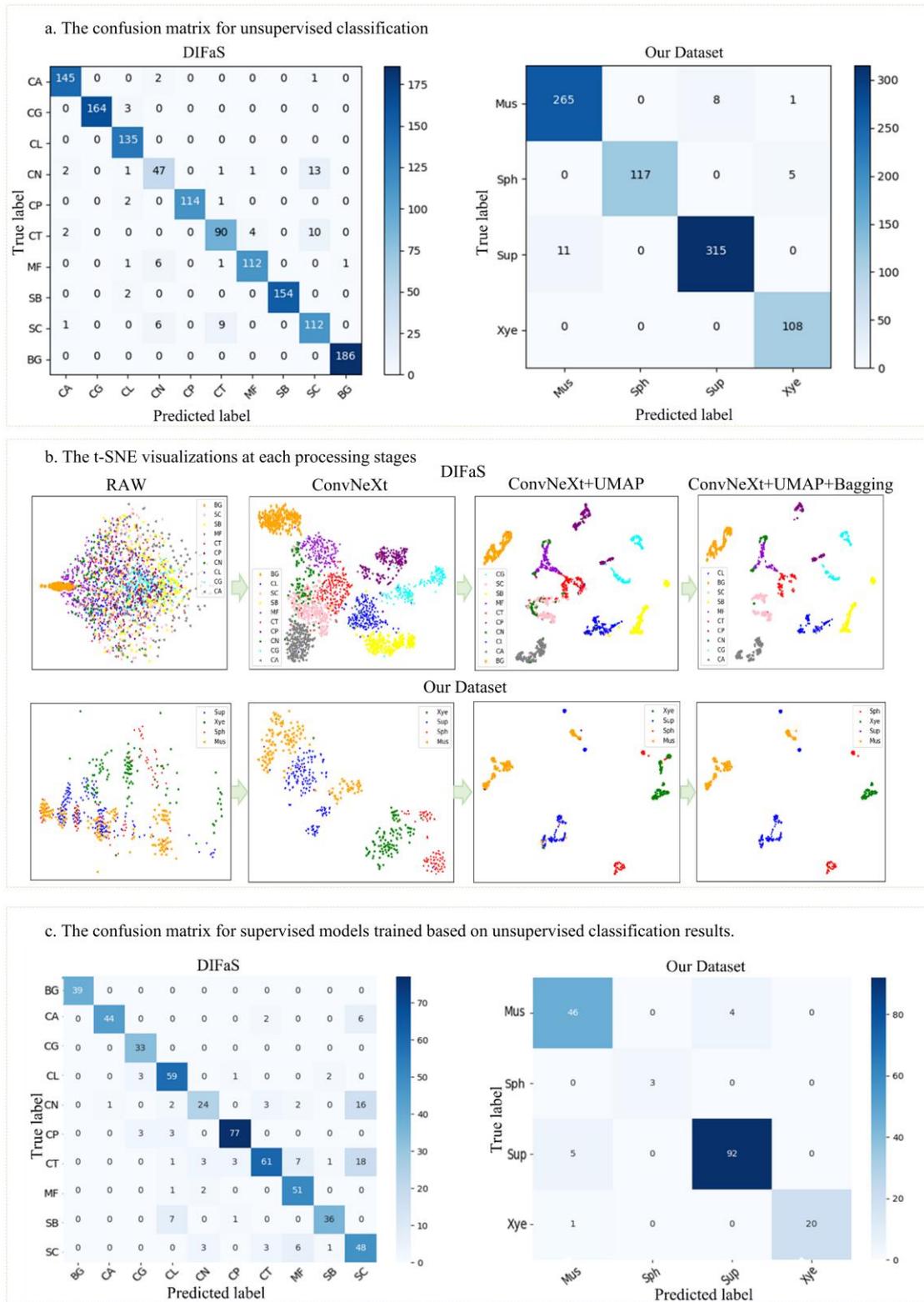

Figure 6. The results presentation. Panel-a. The confusion matrix for unsupervised classification. Panel-b. t-SNE visualization. Panel-c. Confusion matrix for supervised classification.

Table 3. The comparison with existing supervised approaches.

| Method | Dataset | Training counts | Test counts | Overall (%) | Reject (%) |
| --- | --- | --- | --- | --- | --- |
| **Our Method** | **DIFaS** | / | **1359 (out of 1902)** | **94.1±0.6** | **28.5±1.8** |
| AlexNet FV SVM (Zielińsk et al., 2020) | DIFaS | 1513 | 1479 | 82.4±0.2 | / |
| SA-AbMILP (Rymarczyk et al., 2021) | DIFaS | 1513 | 1479 | 86.0±1.0 | / |

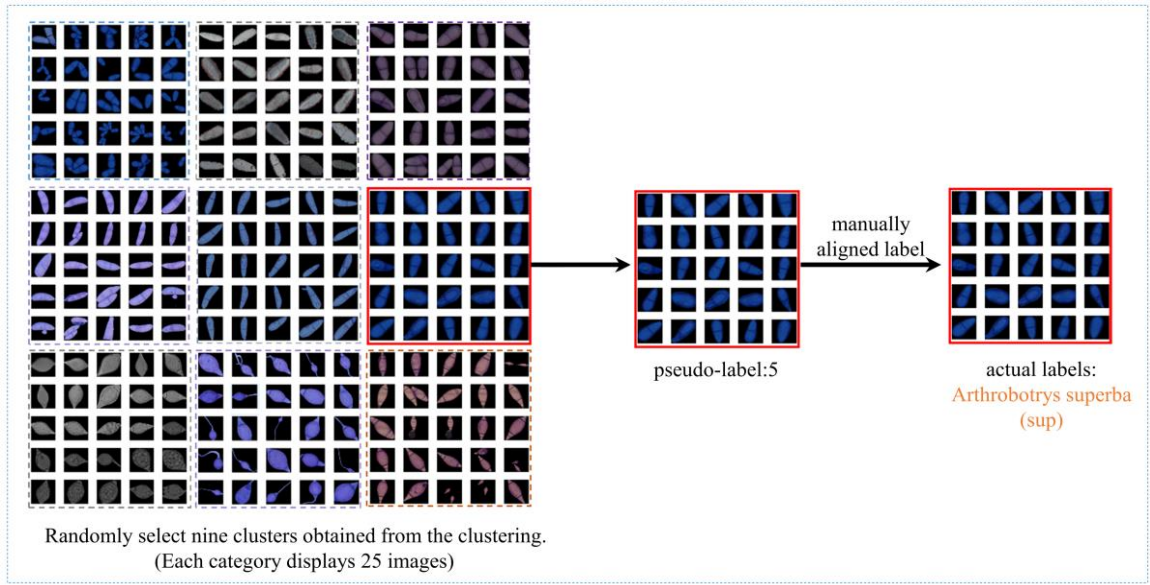

Figure 7. Display of clustering results and the process of manual label alignment (actual clustering consists of 20 classes; here, nine classes are presented).

## 3.2 Train supervised model based on the unsupervised classification results

This section details the outcomes of supervised models that are trained using data classified by our unsupervised approach. Utilizing the classification results from our unsupervised method for training supervised models serves a dual purpose: it allows for a secondary classification of the samples that are initially discarded and expedites the growth of the specialized domain database. To address the challenge of limited data, we employed the pre-trained VGG16 (Simonyan & Zisserman, 2014) for training the supervised model. The data set aside by the voting process is designated as the test set, with specific data divisions illustrated in Table 4. The classification outcomes of the supervised model are presented in Table 5. For the DIFaS dataset, VGG16 achieved an overall accuracy of 82.4%. For the private dataset, the accuracy is 94.2%. To offer a visual assessment of the recognition performance across different classes, we plotted the confusion matrices, as depicted in Figure 6-c.

Table 4. The dataset partitioning for training the supervised model.

|  | Training set size | Test set size |
| --- | --- | --- |
| DIFaS | 1329 | 573 |
| Our Dataset | 830 | 171 |

Table 5. Classification Results Based on Pretrained Supervised Model with VGG16

| DIFaS | CA | CG | CL | CN | CP | CT | MF | SB | SC | BG | Overall(%) |
| --- | --- | --- | --- | --- | --- | --- | --- | --- | --- | --- | --- |
|  | 97.8 | 84.6 | 80.8 | 75.0 | 93.9 | 88.4 | 77.3 | 90.9 | 54.5 | 100.0 | 82.4 |

| Our Dataset | Mus | Sph | Sup | Xye | Overall(%) |
| --- | --- | --- | --- | --- | --- |
|  | 88.5 | 100.0 | 95.8 | 100.0 | 94.2 |

## 3.3 Further analysis

In this section, ablation and comparative experiments are conducted to validate the effectiveness of the proposed method. The performance of ConvNeXt, UMAP, and the voting strategy employed in the multi-clustering method is assessed. The proposed approach is compared with few-shot learning methods. Furthermore, detailed analyses of the impacts of feature alignment and class balance on unsupervised classification are provided in Appendix A1 and A2.

### 3.3.1 Effectiveness of ConvNeXt

Different feature extraction schemes are evaluated, with details provided in Table 6. It is observed that, on both datasets, feature extraction using the ViT and ConvNeXt models yields better clustering accuracy than directly using original images. Notably, the ConvNeXt feature extraction outperforms other methods.

For the private dataset, existing unsupervised feature extracting methods such as the constraints convolutional autoencoder (CCAE) (Hou et al., 2021; Zhu et al., 2022), tensor decomposition methods, such as the CP and Tucker decomposition methods (Rabanser, Shchur, & Günnemann, 2017), are also compared. It shows in Table 6-b that feature extraction with pretrained models proves superior to other methods.

Table 6-a. The comparison of different feature extraction methods on the DIFaS dataset (comparing clustering results obtained using k-means, agglomerative clustering, and Birch after feature extraction).

| Dataset | Pretrained model | Clustering method | Overall accuracy (%) |
|---|---|---|---|
| DIFaS | RAW | kmeans | 41.0±0.8 |
| | | agg | 41.9±0.0 |
| | | birch | 40.8±0.0 |
| | VIT | kmeans | 84.8±1.0 |
| | | agg | 83.9±0.0 |
| | | birch | 83.9±0.0 |
| | **Convnext** | **kmeans** | **84.2±1.0** |
| | | **agg** | **86.8±0.0** |
| | | **birch** | **87.4±0.0** |

Table 6-b. The comparison of different feature extraction methods on the Our dataset (comparing clustering results obtained using k-means, agglomerative clustering, and Birch after feature extraction).

| Dataset | Pretrained model | Clustering method | Overall accuracy (%) |
|---|---|---|---|
| Our Dataset | RAW | kmeans | 50.0±1.0 |
| | | agg | 48.3±0.0 |
| | | birch | 48.3±0.0 |
| | Tucker | kmeans | 52.4±1.6 |
| | | agg | 51.3±0.0 |
| | | birch | 51.3±0.0 |
| | CCAE | kmeans | 50.4±0.9 |
| | | agg | 52.2±0.0 |
| | | birch | 52.2±0.0 |
| | Cp | kmeans | 53.4±0.7 |
| | | agg | 54.0±0.0 |
| | | birch | 54.0±0.0 |
| | VIT | kmeans | 75.1±1.9 |
| | | agg | 76.4±0.0 |
| | | birch | 76.4±0.0 |

|  |  |  |
|---|---|---|
|  | kmeans | 90.9±0.9 |
| Convnext | agg | 93.0±0.0 |
|  | birch | 93.0±0.0 |

### 3.3.2 Effectiveness of UMAP

This section contrasts the linear dimensionality reduction technique, PCA (Maćkiewicz & Ratajczak, 1993), with nonlinear manifold learning methodologies (Izenman, 2012).

Among them, Laplacian Eigenmaps (LE) (Li, Li, & Zhang, 2019) is a method used for nonlinear dimensionality reduction and graph embedding. The core idea of this method is to map data to a low-dimensional feature space through spectral decomposition. Locally Linear Embedding (LLE) (Roweis & Saul, 2000) aims to find the optimal linear representation within local neighborhoods and then reduce dimensionality by minimizing the reconstruction error. Isometric Mapping (Isomap) (Akkucuk & Carroll, 2006) is dedicated to preserving the geodesic distances between data points. It achieves this by constructing a matrix of geodesic distances between data points and then maintaining these distances in a low-dimensional space. The dimension selection for linear dimensionality reduction methods such as PCA follows the "elbow method," while for non-linear dimensionality reduction methods like UMAP, the choice of the optimal target dimension is based on clustering performance. Further details can be found in Appendix A3 and A4. It shows that, across both datasets, UMAP outperforms scenarios with no dimensionality reduction (No DR) and other methodologies, boosting accuracy by 4-5% on the DIFaS dataset, as seen in Table 7-a. On the private dataset, UMAP also demonstrated improvements. Separately, an attempt is made to employ contrastive learning (CL) (Chen et al., 2020; Gao et al., 2023) on the private dataset using grayscale-augmented images as positives. Yet, as indicated in Table 7-b, the classification accuracy is less than the optimal. Although contrastive learning is an effective learning strategy, the setup of positive samples in unsupervised learning can influence the final classification performance. Thus, how to construct positive samples becomes a challenge when utilizing CL strategies for feature extraction (Zhou et al., 2023).

Table 7-a. The comparison of different dimensionality reduction methods on the DIFaS dataset (evaluating the effectiveness of different dimensionality reduction methods based on clustering accuracy).

| Dataset | Pretrained model | Dimension reduction | Clustering method | Overall accuracy (%) |
|---|---|---|---|---|
| DIFaS | Convnext | LE | kmeans | 81.8±1.7 |
| | | | agg | 80.4±0.0 |
| | | | birch | 11.8±0.0 |
| | | LLE | kmeans | 82.8±1.1 |
| | | | agg | 82.1±0.0 |
| | | | birch | 51.7±0.0 |
| | | Isomap | kmeans | 86.1±1.2 |
| | | | agg | 82.6±0.0 |
| | | | birch | 82.2±0.0 |
| | | No DR | kmeans | 84.2±1.0 |
| | | | agg | 86.8±0.0 |
| | | | birch | 87.4±0.0 |
| | | PCA | kmeans | 85.0±0.9 |
| | | | agg | 86.8±0.0 |
| | | | birch | 87.4±0.0 |
| | | **UMAP** | **kmeans** | **91.2±0.3** |
| | | | **agg** | **90.6±0.0** |
| | | | **birch** | **91.4±0.0** |

Table 7-b．The comparison of different dimensionality reduction methods on the Our dataset (evaluating the effectiveness of different dimensionality reduction methods based on clustering accuracy)

| Dataset | Pretrained model | Dimension reduction | Clustering method | Overall accuracy (%) |
|---|---|---|---|---|
| Our Dataset | Convnext | CL | kmeans | 76.9±4.4 |
| | | | agg | 86.4±0.0 |
| | | | birch | 86.4±0.0 |
| | | Isomap | kmeans | 88.0±0.7 |
| | | | agg | 88.6±0.0 |
| | | | birch | 88.6±0.0 |
| | | LE | kmeans | 90.5±0.9 |
| | | | agg | 91.6±0.0 |
| | | | birch | 42.3±0.0 |
| | | LLE | kmeans | 91.7±0.3 |
| | | | agg | 90.7±0.0 |
| | | | birch | 42.3±0.0 |
| | | No DR | kmeans | 90.9±0.9 |

|  |  |  |
|---|---|---|
|  | agg | 93.0±0.0 |
|  | birch | 93.0±0.0 |
|  | kmeans | 91.6±2.1 |
| PCA | agg | 93.0±0.0 |
|  | birch | 93.0±0.0 |
|  | **kmeans** | **92.9±0.7** |
| UMAP | **agg** | **95.2±0.0** |
|  | **birch** | **92.0±0.0** |

### 3.3.3 Effectiveness of voting strategy

In this section, Table 8 showcases the classification results of individual methods against bagging to underscore the effectiveness of the voting strategy. For the DIFaS dataset, bagging enhances accuracy by over 2.7%, albeit with around 28.5% of samples being rejected compared to individual methods. On the private dataset, there's an accuracy improvement of over 1.5%, but about 17.2% of samples are dismissed. These results indicate that ensemble learning fosters more robust and accurate clustering outcomes. Despite the trade-off in sample loss, the high-purity results obtained from ensemble clustering can be directly harnessed as pre-labeled datasets for subsequent tasks.

Table 8-a. Comparing the performance of bagging on the DIFaS dataset

| Method |  | Overall accuracy (%) | Reject (%) |
|---|---|---|---|
|  | kmeans | 91.2±0.3 | 0 |
| Convnext+UMAP | agg | 90.6±0.0 | 0 |
|  | birch | 91.4±0.0 | 0 |
| **Convnext+UMAP +bagging (Our method)** |  | **94.1±0.6** | **28.5±1.8%** |

Table 8-b. Comparing the performance of bagging on the private dataset

| Method |  | Overall accuracy (%) | Reject (%) |
|---|---|---|---|
|  | kmeans | 92.9±0.7 | 0 |
| Convnext+UMAP | agg | 95.2±0.0 | 0 |
|  | birch | 92.0±0.0 | 0 |
| **Convnext+UMAP +bagging (Our method)** |  | **96.7±0.2** | **17.2±2.5** |

### 3.3.4 Exploration of the number of clusters

The number of clusters is a key issue in unsupervised clustering. Too many clusters would increase the pressure of manual annotation, while too few would degrade clustering purity. We explored the impact of different cluster numbers on the private dataset (actual number is 4 classes), as shown in Table 9. The results demonstrate that higher cluster numbers lead to higher accuracy. 20 clusters give the optimal results, ensuring relatively high accuracy without bringing excessive burden for subsequent manual annotation. Our experiments show that it takes only 2 minutes for one person to quickly annotate the 20-cluster sample set. In summary, by properly configuring the clustering granularity, a balance can be achieved between accuracy and labeling efficiency. Reasonable clustering strategies can help generate high-quality pre-labeled datasets and reduce manual annotation workload.

Table 9. Classification Results for Different Cluster Numbers

| Method | | Overall accuracy (%) | Reject (%) |
| --- | --- | --- | --- |
| Our method (Cluster8) | kmeans | 79.7±0.1 | 0 |
| | agg | 79.1±0.0 | 0 |
| | birch | 82.5±0.0 | 0 |
| | bagging | 85.7±0.1 | 16.5±0.5 |
| Our method(Cluster12) | kmeans | 90.3±0.1 | 0 |
| | agg | 90.4±0.0 | 0 |
| | birch | 90.4±0.0 | 0 |
| | bagging | 91.5±0.0 | 3.9±0.0 |
| Our method(Cluster16) | kmeans | 92.9±0.8 | 0 |
| | agg | 94.0±0.0 | 0 |
| | birch | 92.0±0.0 | 0 |
| | bagging | 95.1±0.3 | 15.0±2.6 |
| **Our method(Cluster20)** | **kmeans** | **92.9±0.7** | **0** |
| | **agg** | **95.2±0.0** | **0** |
| | **birch** | **92.0±0.0** | **0** |
| | **bagging** | **96.7±0.2** | **17.2±2.5** |

### 3.3.5 Comparison with few-shot learning

In this section, feature extraction is performed on the private dataset based on pretrained models and UMAP. Classifiers are trained using a few-shot learning approach, and their classification performance is evaluated. Specifically, three machine learning methods SVM (Suthaharan & Suthaharan, 2016), RF (Breiman, 2001) and MLP (Zhou et al.,

2020; Kruse et al., 2022) are used. The results are shown in Table 10. It can be seen that as the number of training samples increases, the classification accuracy of the models gradually improves. The highest classification accuracy is 95.7% given by RF, with 300 labeled samples used for training. Notably, this is still lower than the 96.7% of our unsupervised classification method.

Table 10. The results of few-shot learning based on feature extraction encoding

|  | 4-Way 5-Shot (%) | 4-Way 25-Shot (%) | 4-Way 50-Shot (%) | 4-Way 75-Shot (%) |
| --- | --- | --- | --- | --- |
| SVM | 67.5±1.3 | 77.2±3.0 | 81.4±5.9 | 92.2±1.4 |
| RF | 84.2±1.7 | 93.9±1.4 | 95.0±1.0 | 95.7±0.8 |
| MLP | 79.9±0.2 | 88.4±7.9 | 85.5±13.1 | 74.8±9.8 |

# 4. Conclusion and discussions

## 4.1 Conclusion

In this work, we introduce a well-conceived unsupervised learning strategy. 1) A dual-step feature dimensionality reduction is employed: initially extracting features from individual image data using pretrained large models, then further eliminating redundant features and accentuating distinct ones through manifold learning based on inter-sample differences. 2) The strategy harnesses voting from multiple distinct clustering algorithms to elevate clustering accuracy. 3) Post-hoc visual label alignment is used in lieu of pre-annotated data.

As demonstrated on fungal data—a representative of specialized domain image data—despite leveraging out-of-box techniques, this strategy successfully overcomes the shortcomings of the limited accuracy of individual clustering methods. It further diminishes the reliance of self-supervised classification methods on manually pre-annotated data, culminating in an efficient, cost-effective closed-loop system for the automated classification of professional domain image data. This holds significant implications for the automated analysis of specialized domain imagery. The simplicity and ease of use of this method will also bring convenience to researchers in various fields in building datasets, promoting AI applications for images in specialized domains.

For example, we are currently applying this method to the automated analysis of specialized domain images such as aquatic plants, astronomical, and ultrasonic muscle images.

## 4.2 Discussions

There are several considerations associated with this method that merit discussion.

1) The feature extraction is dependent on pretrained models tailored to general image distributions. Their capability might be compromised when faced with specialized or unseen images, such as ultrasound scans. Similarly, for the automatic annotation of non-image data, we need to construct an appropriate feature extraction module. This is combined with manifold dimensionality reduction to eliminate redundant information through sample comparison and retain differentiated information for unsupervised feature extraction. For instance, in our preliminary work (Gao et al., 2023), we designed a feature extraction module for the combination of enzymes and substrates, and achieved high-precision classification of samples without pre-labeled experimental data.

2) Alignment of images in the feature space is crucial. As demonstrated in Appendix A1, better and more consistent outcomes are achieved by UMAP and other methodologies when orientations of the fungal data are harmonized. This observation suggests the potential benefits of synchronously optimizing feature extraction and manifold learning, paving the way for further enhanced classification performance. Consequently, an end-to-end integration, which directly finetunes feature learning via a UMAP loss, is under exploration.

3) Beyond relying solely on discriminative models for data representation, the utilization of generative models could supplement and boost the existing unsupervised classification framework.

Moreover, 4) as detailed in Appendix A4, determining the optimal target dimension size during UMAP-based dimensionality reduction is also an avenue that requires a robust strategy. In our future work, we will address those mentioned issues of identifying the optimal target dimension of UMAP.

# Acknowledgements

This work was supported by National Nature Science Foundation of China (62106033,42367066,32360002).

# Author contributions statement

Conceptualization, Chichun Zhou; Carrying out the experimental work, Zhaocong Liu and Lin Cheng; Data preparation, Fa Zhang, Huanxi Deng, and Xiaoyan Yang; Formal analysis, Chichun Zhou and Zhenyu; Writing original draft preparation, Zhaocong Liu, Fa Zhang, and Chichun Zhou; Analyzing the results, Zhaocong Liu, Lin Cheng, Xiaoyan Yang and Chichun Zhou; Reviewing the manuscript, Chichun Zhou and Zhenyu Zhang. All authors have read and agreed to the published version of the manuscript.

# Additional information

Conflicts of Interest: The authors declare that they have no known competing financial interests or personal relationships that could have appeared to influence the work reported in this paper.

# References

Ackermann, M. R., Blömer, J., Kuntze, D., & Sohler, C. (2014). Analysis of agglomerative clustering. Algorithmica, 69, 184-215.

Ajay, P., Nagaraj, B., Kumar, R. A., Huang, R., & Ananthi, P. (2022). Unsupervised hyperspectral microscopic image segmentation using deep embedded clustering algorithm. Scanning, 2022.

Akkucuk, U., & Carroll, J. D. (2006). PARAMAP vs. Isomap: a comparison of two nonlinear mapping algorithms. Journal of Classification, 23, 221-254.

Albelwi, S. (2022). Survey on self-supervised learning: auxiliary pretext tasks and contrastive learning methods in imaging. Entropy, 24(4), 551.

Breiman, L. (2001). Random forests. Machine learning, 45, 5-32.

Carion, N., Massa, F., Synnaeve, G., Usunier, N., Kirillov, A., & Zagoruyko, S. (2020, August). End-to-end object detection with transformers. In European conference on computer vision (pp. 213-229). Cham: Springer International Publishing.

Caron, M., Misra, I., Mairal, J., Goyal, P., Bojanowski, P., & Joulin, A. (2020). Unsupervised learning of visual features by contrasting cluster assignments. Advances in neural information processing systems, 33, 9912-9924.

Chen, D., Chen, Y., Li, Y., Mao, F., He, Y., & Xue, H. (2021, June). Self-supervised learning for few-shot image classification. In ICASSP 2021-2021 IEEE International Conference on Acoustics, Speech and Signal Processing (ICASSP) (pp. 1745-1749). IEEE.

Chen, T., Kornblith, S., Norouzi, M., & Hinton, G. (2020, November). A simple framework for contrastive learning of visual representations. In International conference on machine learning (pp. 1597-1607). PMLR.

Conneau, A., Schwenk, H., Barrault, L., & Lecun, Y. (2016). Very deep convolutional networks for text classification. arXiv preprint arXiv:1606.01781.

Dai, Y., Xu, J., Song, J., Fang, G., Zhou, C., Ba, S., ... & Kong, X. (2023). The Classification of Galaxy Morphology in the H Band of the COSMOS-DASH Field: A Combination-based Machine-learning Clustering Model. *The Astrophysical Journal Supplement Series*, *268*(1), 34.

Deng, J., Dong, W., Socher, R., Li, L. J., Li, K., & Fei-Fei, L. (2009, June). Imagenet: A large-scale hierarchical image database. In 2009 IEEE conference on computer vision and pattern recognition (pp. 248-255). Ieee.

Dey, A., Schlegel, D. J., Lang, D., Blum, R., Burleigh, K., Fan, X., ... & Vivas, A. K. (2019). Overview of the DESI legacy imaging surveys. The Astronomical Journal, 157(5), 168.

Ericsson, L., Gouk, H., Loy, C. C., & Hospedales, T. M. (2022). Self-supervised representation learning: Introduction, advances, and challenges. IEEE Signal Processing Magazine, 39(3), 42-62.

Fang, G., Ba, S., Gu, Y., Lin, Z., Hou, Y., Qin, C., ... & Kong, X. (2023). Automatic classification of galaxy morphology: A rotationally-invariant supervised machine-learning method based on the unsupervised machine-learning data set. The Astronomical Journal, 165(2), 35.

Gao, L., Yu, Z., Wang, S., Hou, Y., Zhang, S., Zhou, C., & Wu, X. (2023). A new paradigm in lignocellulolytic enzyme cocktail optimization: Free from expert-level prior knowledge and experimental datasets. Bioresource Technology, 129758.

Goodfellow, I., Pouget-Abadie, J., Mirza, M., Xu, B., Warde-Farley, D., Ozair, S., ... & Bengio, Y. (2014). Generative adversarial nets. Advances in neural information processing systems, 27.

Hartigan, J. A., & Wong, M. A. (1979). Algorithm AS 136: A k-means clustering algorithm. Journal of the royal statistical society. series c (applied statistics), 28(1), 100-108.

He, K., Zhang, X., Ren, S., & Sun, J. (2016). Deep residual learning for image recognition. In Proceedings of the IEEE conference on computer vision and pattern recognition (pp. 770-778).

Hou, Y. J., Xie, Z. X., & Zhou, C. C. (2021). An unsupervised deep-learning method for fingerprint classification: the ccae network and the hybrid clustering strategy. arXiv preprint arXiv:2109.05526.

Huda, W., & Abrahams, R. B. (2015). X-ray-based medical imaging and resolution. American Journal of Roentgenology, 204(4), W393-W397.

Izenman, A. J. (2012). Introduction to manifold learning. Wiley Interdisciplinary Reviews: Computational Statistics, 4(5), 439-446.

Jing, L., & Tian, Y. (2020). Self-supervised visual feature learning with deep neural networks: A survey. IEEE transactions on pattern analysis and machine intelligence, 43(11), 4037-4058.

Kruse, R., Mostaghim, S., Borgelt, C., Braune, C., & Steinbrecher, M. (2022). Multi-layer perceptrons. In Computational intelligence: a methodological introduction (pp. 53-124). Cham: Springer International Publishing.

Kuznetsova, A., Rom, H., Alldrin, N., Uijlings, J., Krasin, I., Pont-Tuset, J., ... & Ferrari, V. (2020). The open images dataset v4: Unified image classification, object detection, and visual relationship detection at scale. International Journal of Computer Vision, 128(7), 1956-1981.

Kadam, S., & Vaidya, V. (2020). Review and analysis of zero, one and few shot learning approaches. In Intelligent Systems Design and Applications: 18th International Conference on Intelligent Systems Design and Applications (ISDA 2018) held in Vellore, India, December 6-8, 2018, Volume 1 (pp. 100-112). Springer International Publishing.

Karras, T., Laine, S., Aittala, M., Hellsten, J., Lehtinen, J., & Aila, T. (2020). Analyzing and improving the image quality of stylegan. In Proceedings of the IEEE/CVF conference on computer vision and pattern recognition (pp. 8110-8119).

Kirillov, A., Mintun, E., Ravi, N., Mao, H., Rolland, C., Gustafson, L., ... & Girshick, R. (2023). Segment anything. arXiv preprint arXiv:2304.02643.

Kumar, P., Rawat, P., & Chauhan, S. (2022). Contrastive self-supervised learning: review, progress, challenges and future research directions. International Journal of Multimedia Information Retrieval, 11(4), 461-488.

Le-Khac, P. H., Healy, G., & Smeaton, A. F. (2020). Contrastive representation learning: A framework and review. Ieee Access, 8, 193907-193934.

Li, B., Li, Y. R., & Zhang, X. L. (2019). A survey on Laplacian eigenmaps based manifold learning methods. Neurocomputing, 335, 336-351.

Lin, T. Y., Maire, M., Belongie, S., Hays, J., Perona, P., Ramanan, D., ... & Zitnick, C. L. (2014). Microsoft coco: Common objects in context. In *Computer Vision–ECCV 2014: 13th European Conference, Zurich, Switzerland, September 6-12, 2014, Proceedings, Part V 13* (pp. 740-755). Springer International Publishing.

Litjens, G., Kooi, T., Bejnordi, B. E., Setio, A. A. A., Ciompi, F., Ghafoorian, M., ... & Sánchez,

C. I. (2017). A survey on deep learning in medical image analysis. Medical image analysis, 42, 60-88.

Liu, S., Wang, Y., Yang, X., Lei, B., Liu, L., Li, S. X., ... & Wang, T. (2019). Deep learning in medical ultrasound analysis: a review. Engineering, 5(2), 261-275.

Liu, X., Zhang, F., Hou, Z., Mian, L., Wang, Z., Zhang, J., & Tang, J. (2021a). Self-supervised learning: Generative or contrastive. IEEE transactions on knowledge and data engineering, 35(1), 857-876.

Liu, Z., Lin, Y., Cao, Y., Hu, H., Wei, Y., Zhang, Z., ... & Guo, B. (2021b). Swin transformer: Hierarchical vision transformer using shifted windows. In Proceedings of the IEEE/CVF international conference on computer vision (pp. 10012-10022).

Liu, Z., Mao, H., Wu, C. Y., Feichtenhofer, C., Darrell, T., & Xie, S. (2022). A convnet for the 2020s. In Proceedings of the IEEE/CVF conference on computer vision and pattern recognition (pp. 11976-11986).

Long, J., Shelhamer, E., & Darrell, T. (2015). Fully convolutional networks for semantic segmentation. In Proceedings of the IEEE conference on computer vision and pattern recognition (pp. 3431-3440).

Lintott, C. J., Schawinski, K., Slosar, A., Land, K., Bamford, S., Thomas, D., ... & Vandenberg, J. (2008). Galaxy Zoo: morphologies derived from visual inspection of galaxies from the Sloan Digital Sky Survey. Monthly Notices of the Royal Astronomical Society, 389(3), 1179-1189.

Luo, Z. L., Hyde, K. D., Liu, J. K., Maharachchikumbura, S. S., Jeewon, R., Bao, D. F., ... & Su, H. Y. (2019). Freshwater sordariomycetes. Fungal diversity, 99, 451-660.

Madsen, J. D., & Wersal, R. M. (2017). A review of aquatic plant monitoring and assessment methods. Journal of Aquatic Plant Management, 55(1), 1-12.

Maćkiewicz, A., & Ratajczak, W. (1993). Principal components analysis (PCA). Computers & Geosciences, 19(3), 303-342.

McInnes, L., Healy, J., & Melville, J. (2018). Umap: Uniform manifold approximation and projection for dimension reduction. arXiv preprint arXiv:1802.03426.

Niu, S., Liu, Y., Wang, J., & Song, H. (2020). A decade survey of transfer learning (2010–2020). IEEE Transactions on Artificial Intelligence, 1(2), 151-166.

Ohri, K., & Kumar, M. (2021). Review on self-supervised image recognition using deep neural networks. Knowledge-Based Systems, 224, 107090.

Ozbulak, U., Lee, H. J., Boga, B., Anzaku, E. T., Park, H., Van Messem, A., ... & Vankerschaver, J. (2023). Know Your Self-supervised Learning: A Survey on Image-based Generative and Discriminative Training. arXiv preprint arXiv:2305.13689.

Paul, S., & Chen, P. Y. (2022, June). Vision transformers are robust learners. In Proceedings of the AAAI conference on Artificial Intelligence (Vol. 36, No. 2, pp. 2071-2081).

Rahman, S., Khan, S., & Porikli, F. (2018). A unified approach for conventional zero-shot, generalized zero-shot, and few-shot learning. IEEE Transactions on Image Processing,

27(11), 5652-5667.

Rabanser, S., Shchur, O., & Günnemann, S. (2017). Introduction to tensor decompositions and their applications in machine learning. arXiv preprint arXiv:1711.10781.

Roweis, S. T., & Saul, L. K. (2000). Nonlinear dimensionality reduction by locally linear embedding. Science, 290(5500), 2323-2326.

Rymarczyk, D., Borowa, A., Tabor, J., & Zielinski, B. (2021). Kernel self-attention for weakly-supervised image classification using deep multiple instance learning. In *Proceedings of the IEEE/CVF Winter Conference on Applications of Computer Vision* (pp. 1721-1730).

Schmarje, L., Santarossa, M., Schröder, S. M., & Koch, R. (2021). A survey on semi-, self- and unsupervised learning for image classification. IEEE Access, 9, 82146-82168.

Simonyan, K., & Zisserman, A. (2014). Very deep convolutional networks for large-scale image recognition. arXiv preprint arXiv:1409.1556.

Suthaharan, S., & Suthaharan, S. (2016). Support vector machine. Machine learning models and algorithms for big data classification: thinking with examples for effective learning, 207-235.

Tan, M., & Le, Q. (2019, May). Efficientnet: Rethinking model scaling for convolutional neural networks. In International conference on machine learning (pp. 6105-6114). PMLR.

Van der Maaten, L., & Hinton, G. (2008). Visualizing data using t-SNE. Journal of Machine Learning Research, 9(11).

Van Gansbeke, W., Vandenhende, S., Georgoulis, S., Proesmans, M., & Van Gool, L. (2020, August). Scan: Learning to classify images without labels. In European conference on computer vision (pp. 268-285). Cham: Springer International Publishing.

Yang, H., Li, C., Zhao, X., Cai, B., Zhang, J., Ma, P., ... & Grzegorzek, M. (2023). EMDS-7: Environmental microorganism image dataset seventh version for multiple object detection evaluation. Frontiers in Microbiology, 14, 1084312.

Yang, J., & Zhang, K. Q. (2014). Biological control of plant-parasitic nematodes by nematophagous fungi. Nematode-trapping fungi, 231-262.

York, D. G., Adelman, J., Anderson Jr, J. E., Anderson, S. F., Annis, J., Bahcall, N. A., ... & Yasuda, N. (2000). The Sloan Digital Sky Survey: Technical summary. The Astronomical Journal, 120(3), 1579.

Zhang, F., Yang, Y. Q., Zhou, F. P., Xiao, W., Boonmee, S., & Yang, X. Y. (2023). Morphological and Phylogenetic Characterization of Five Novel Nematode-Trapping Fungi (Orbiliomycetes) from Yunnan, China. Journal of Fungi, 9(7), 735.

Zhang, K. Q., & Mo, M. H. (2006). Flora fungorum sinicorum, vol. 33. Arthrobotrys et genera cetera cognata.

Zhang, L., & Du, B. (2012). Recent advances in hyperspectral image processing. Geo-spatial Information Science, 15(3), 143-156.

Zhang, T., Ramakrishnan, R., & Livny, M. (1996). BIRCH: an efficient data clustering method for very large databases. ACM SIGMOD Record, 25(2), 103-114.

Zhang, Y., Li, W., Zhang, M., & Tao, R. (2022, May). Dual graph cross-domain few-shot learning for hyperspectral image classification. In ICASSP 2022-2022 IEEE International Conference on Acoustics, Speech and Signal Processing (ICASSP) (pp. 3573-3577). IEEE.

Zhou, C., Gu, Y., Fang, G., & Lin, Z. (2022). Automatic morphological classification of galaxies: convolutional autoencoder and bagging-based multiclustering model. The Astronomical Journal, 163(2), 86.

Zhou, C., Guan, X., Yu, Z., Shen, Y., Zhang, Z., & Gu, J. (2023). An Innovative Unsupervised Gait Recognition Based Tracking System for Safeguarding Large-Scale Nature Reserves in Complex Terrain. Available at SSRN 4598768.

Zhou, C., Tu, H., Hou, Y., Ling, Z., Liu, Y., & Hua, J. (2020). Activation functions are not needed: the ratio net. arXiv preprint arXiv:2005.06678.

Zhu, H., Nie, W. J., Hou, Y. J., Du, Q. M., Li, S. J., & Zhou, C. C. (2022). An Unsupervised Deep-Learning Method for Bone Age Assessment. arXiv preprint arXiv:2206.05641.

Zieliński, B., Sroka-Oleksiak, A., Rymarczyk, D., Piekarczyk, A., & Brzychczy-Włoch, M. (2020). Deep learning approach to describe and classify fungi microscopic images. PloS one, 15(6), e0234806.

# Appendix

## A1. The Impact of image feature alignment on classification performance

For the private fungal images, another preprocessing strategy is also applied, where conid iospore images are manually cropped using labelme. Considering that the extracted fungal samples may have different orientations, to prevent the model from learning orientation as a feature and thus avoid bias in the test results, we rotate all samples to the same orientation to align their spatial features, as shown in Figure a-1. After this processing, the number of samples obtained is shown in Table a-1, and the results of our method are presented in Table a-2. The analysis shows that clearer samples and spatial feature alignment help further improve classification accuracy. Also, under such spatial feature alignment, UMAP leads to larger improvement in classification accuracy, as shown in Table a-3, with about 9% increase for k-means clustering.

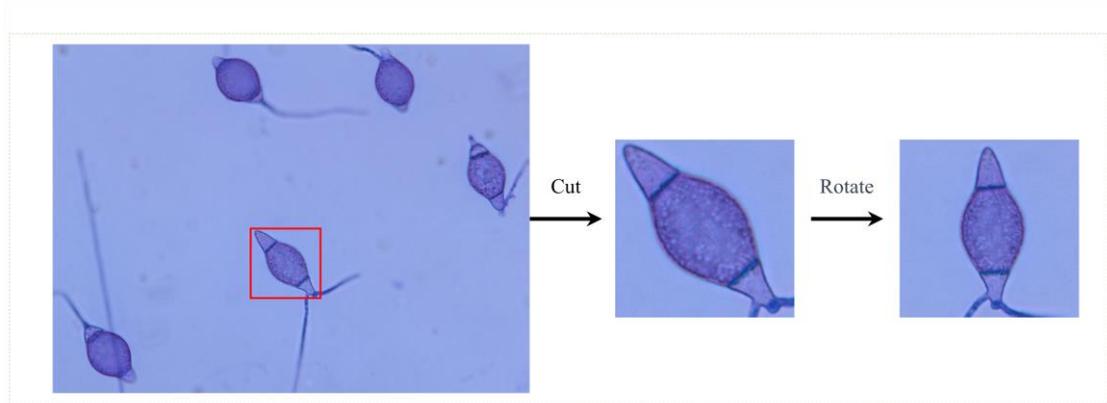

Figure a-1. Preprocessing Schematic

Table a-1. Sample Descriptions

| Our Dataset (a) | Category | Mus | Sph | Sup | Xye | Total |
|---|---|---|---|---|---|---|
| | Counts | 155 | 86 | 154 | 108 | 503 |

Table a-2. Overall Classification Results

| Our Dataset (a) | Mus | Sph | Sup | Xye | Overall(%) | Reject(%) |
|---|---|---|---|---|---|---|
| Convnext+UMAP+bagging (Our Method) | 96.2±0.0 | 100.0±0.0 | 96.7±0.0 | 100.0±0.0 | 97.6±0.0 | 10.3±0.0% |

Table a-3. Enhancement of UMAP Effect

| Feature extraction | Clustering | Overall (%) |
|---|---|---|
| Convnext | kmeans | 87.3±4.0 |
| | agg | 91.5±0.0 |
| | birch | 91.5±0.0 |
| **Convnext+UMAP** | **kmeans** | **96.2±0.0** |
| | **agg** | **96.4±0.0** |
| | **birch** | **91.3±0.0** |

## A2. The impact of imbalance on classification performance

To analyze the impact of sample class distribution on classification performance, balanced and imbalanced versions of the private dataset are constructed, as shown in Table a-4. The experimental results in Table a-5 show that compared to imbalanced data, balanced data has slightly higher classification accuracy and fewer discarded samples. Overall, the distribution of sample classes affects the performance of unsupervised learning. However, unsupervised learning is more robust to long-tailed distributions compared to supervised learning. Future research could explore how to automatically assess class distribution and perform adaptive sampling.

Table a-4. Balanced Data and Imbalanced Data Samples

|  | Category | Mus | Sph | Sup | Xye | Total |
|---|---|---|---|---|---|---|
| Our Dataset (a) | Counts | 155 | 86 | 154 | 108 | 503 |
| Our Dataset (b) | Counts | 80 | 80 | 80 | 80 | 320 |

Table a-5. Comparison of Accuracy Between Balanced and Imbalanced Data Based on Our Method

|  | Dataset | Mus | Sph | Sup | Xye | Overall | Reject(%) |
|---|---|---|---|---|---|---|---|
| Our Method | Our Dataset (a) | 96.2±0.0 | 100.0±0.0 | 96.7±0.0 | 100.0±0.0 | 97.6±0.0 | 10.3±0.0% |
|  | Our Dataset (b) | 96.2±0.1 | 98.7±0.0 | 100.0±0.0 | 100.0±0.0 | 98.7±0.0 | 4.6±0.5% |

## A3. Dimension selection for linear dimensionality reduction methods

In linear dimensionality reduction, choosing an appropriate number of dimensions is a key issue. It is generally assessed by explaining variance or information loss. linear dimensionality reduction, choosing an appropriate number of dimensions is a key issue. It is generally assessed by explaining variance or information loss. We adopt the "elbow method": calculate the variance of each principal component, observe the information ratio curve (information $ratio = \text{eigenvalue}_i/\text{eigenvalue}_{i+1}$) and find the "elbow point". When the ratio is close to 1, increasing dimensions retains more information. When the ratio drops rapidly, the next principal component no longer provides significant information. As shown in Figure a-2, after feature extraction the dimension is 2048 for the private dataset. Comprehensively looking at the change of information ratio, around 1250 appears to be the "elbow point", indicating 1250 as a suitable reduced dimension.

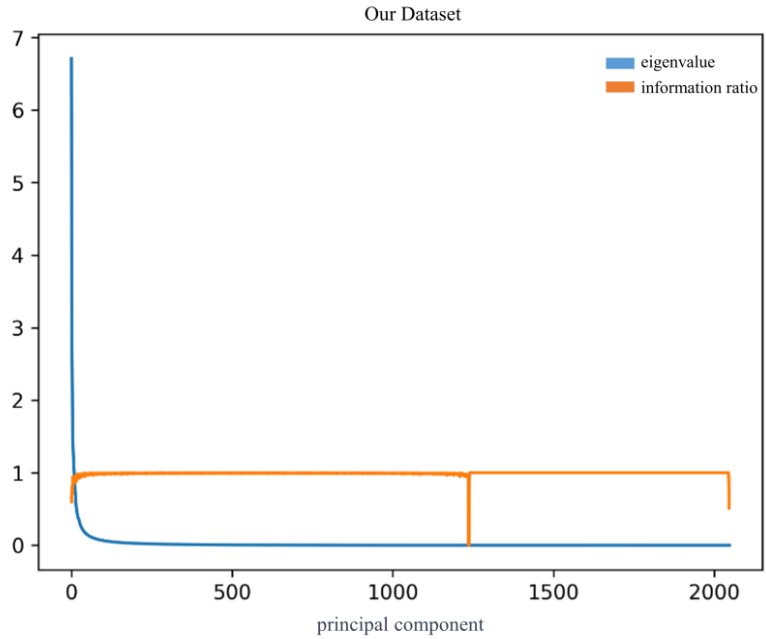

Figure a-2. Principal Component Eigenvalue

## A4. UMAP dimension reduction selection

When using UMAP for nonlinear dimensionality reduction, choosing an appropriate dimension for the low-dimensional space is crucial for performance. The optimal target dimension is selected based on clustering performance. Specifically, taking the private dataset as an example, the original data has 2048 dimensions after feature extraction by the pretrained large model. The clustering performance after reduction to 50, 100, 200, 500 dimensions with UMAP is compared, as shown in Figure a-3. The results show that the highest and most stable classification accuracy can be obtained at 200 dims. Through comprehensive analysis, it is determined that the optimal UMAP target dimension is 200 dims, and it is chosen accordingly to optimize clustering performance. The target dimension selection for other manifold learning methods is also based on clustering performance.

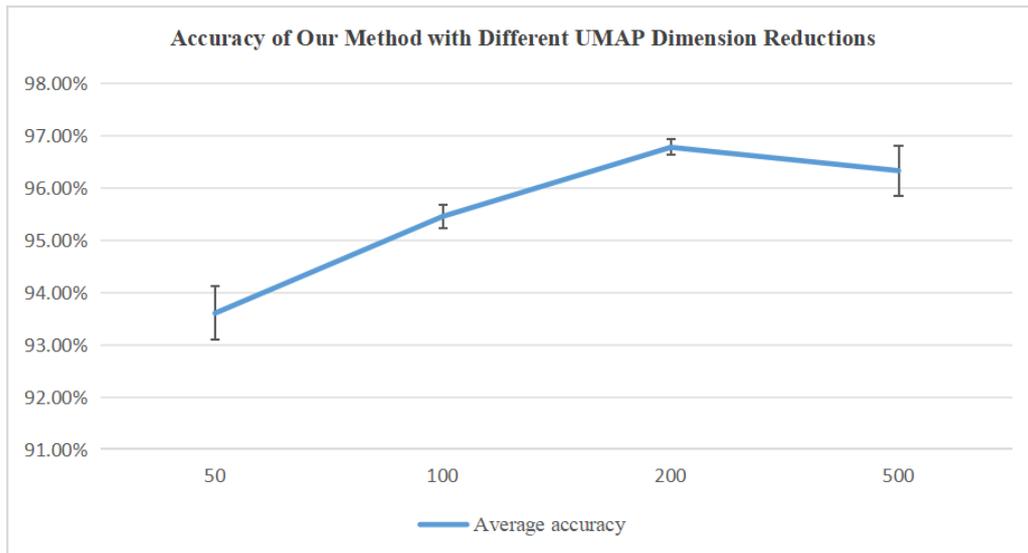

Figure a-3. Accuracies with different UMAP dimension reductions